\documentclass{article}

\usepackage{PRIMEarxiv}

\usepackage[utf8]{inputenc} 
\usepackage[T1]{fontenc}    
\usepackage{hyperref}       
\usepackage{url}            
\usepackage{booktabs}       
\usepackage{multirow}
\usepackage{amsfonts}       
\usepackage{nicefrac}       
\usepackage{microtype}      
\usepackage{lipsum}
\usepackage{fancyhdr}       
\usepackage{graphicx}       
\graphicspath{{media/}}     
\usepackage{enumerate}

\usepackage{amsmath}
\usepackage{amsthm}
\usepackage{algorithm}
\usepackage{algorithmic}

\usepackage{bm}
\usepackage{subfigure}

\usepackage[numbers,sort&compress]{natbib}

\usepackage{xcolor}

\newcommand{\abs}[1]{\left\lvert#1\right\rvert}

\newcommand{\bmf}[1]{\boldsymbol#1}

\usepackage{fourier} 
\usepackage{array}
\usepackage{makecell}

\usepackage{xspace}
\newcommand{\fedavg}{\texttt{FedAvg}\xspace}
\newcommand{\fedprox}{\texttt{FedProx}\xspace}
\newcommand{\AMS}{\texttt{AMS}\xspace}
\newcommand{\PFNM}{\texttt{PFNM}\xspace}
\newcommand{\AMSfull}{\texttt{AMS-full}\xspace}
\newcommand{\AMStop}{\texttt{AMS-top1}\xspace}
\title{Investigating Neuron Disturbing in Fusing Heterogeneous Neural Networks
}

\author{
  Biao~Zhang \\
  School of Mathematical Sciences \\
  Fudan University \\
  Shanghai, China\\
  \texttt{zhangb20@fudan.edu.cn} \\
  \And
  Shuqin~Zhang \thanks{Corresponding author} \\
  School of Mathematical Sciences\\
  Fudan University\\
  Shanghai, China\\
  \texttt{zhangs@fudan.edu.cn}\\
}

\begin{document}
\maketitle

\large

\begin{abstract}
Fusing deep learning models trained on separately located clients into a global model in a one-shot communication round is a straightforward implementation of Federated Learning. Although current model fusion methods are shown experimentally valid in fusing neural networks with almost identical architectures, they are rarely theoretically analyzed. In this paper, we reveal the phenomenon of neuron disturbing, where neurons from heterogeneous local models interfere with each other mutually. We give detailed explanations from a Bayesian viewpoint combining the data heterogeneity among clients and properties of neural networks. Furthermore, to validate our findings, we propose an experimental method that excludes neuron disturbing and fuses neural networks via adaptively selecting a local model, called \AMS, to execute the prediction according to the input. The experiments demonstrate that \AMS is more robust in data heterogeneity than general model fusion and ensemble methods. This implies the necessity of considering neural disturbing in model fusion.  Besides, \AMS is available for fusing models with varying architectures as an experimental algorithm, and we also list several possible extensions of \AMS for future work. 
\end{abstract}


\section{Introduction}
\label{sec:introduction}
\par As concerns of privacy protection grow recently, federated learning algorithms \citep{mcmahan2017communication, li2020federated}, in which the model is learned without data transmission among clients, develop rapidly. Communication costs between clients and the server, and client heterogeneity are two key challenges in federated training. As the communication frequency, rounds of communication during a certain period of time must increase to alleviate the discrepancy among the learned model on clients. Classic algorithms such as \fedavg \cite{mcmahan2017communication}, and \fedprox \cite{li2018federated} focus on training a model via multiple communication rounds of parameters, which results in high privacy risk and communication cost.  Different from this paradigm, model fusion methods are designed for fusing those local models from clients into a global model \cite{utans1996weight,  yurochkin2019bayesian, yurochkin2019statistical,yurochkin2018scalable, wang2020federated,singh2020model, claici2020model,lin2020ensemble,zhu2021data,chen2020fedbe, liu2022deep, xiao2023probabilistic} in a one-shot manner based only on the weights of the local models. 

\par The anterior model fusion studies \cite{yurochkin2019bayesian, wang2020federated, singh2020model} points out that the directly average of parameters is not rational due to the permutation invariant property of weights, and the channels from different networks are always randomly permuted. Thus, many algorithms formulate the fusing problem into alignment problem, including linear assignment \cite{yurochkin2019bayesian, wang2020federated}, optimal transport \cite{singh2020model, nguyen2021model} and graph matching \cite{liu2022deep}. These researches assume that all local neural networks share the same architectures, though a varying number of neurons in each layer is allowed. Besides, a fundamental implicit assumption shared by them is,  if the probability measurement of parameters in the fused global model well approximates the probability measurement of neurons in local models in a specific sampling manner, then the performance of those local models is also maintained (the neuron approximation assumption, NA assumption). This assumption acknowledges the feasibility of model fusion. Nevertheless, the rationality of this prior knowledge under federated learning settings has not been thoroughly investigated. In this paper, we reveal that when the data heterogeneity and model optimization procedure discrepancy among clients are large, the above assumption generally does not hold because of neuron disturbing, in which neurons extracted from heterogeneous clients (data distributions on clients are non-IID and model optimization procedures are different) disturbs each other and harms the fused model performance.  We present a basic analysis of neuron disturbing from a Bayesian view. Furthermore, inspired by a phenomenon among heterogeneous models we called "absolute confidence of neural networks", we propose model fusion via adaptively selecting a local model (\AMS). As an experimental algorithm,  \AMS shows robustly better performance on fusing multilayer perceptron neural networks (MLPs) and convolutional neural networks (CNNs) trained with datasets with two kinds of data partitions. The performance of the other methods including the ensemble method, \fedavg, and \PFNM declines rapidly when the severity of data heterogeneity reaches some poi. Those results verify the existence of neuron disturbing and indicate the necessity of handling neuron disturbing in developing model fusion methods. Besides, in light of computational complexity, we also list possible extensions of \AMS for real-world applications not limited to federated learning. 

\section{Related Work}
\subsection{Federated Learning}
Federated learning aims to learn a shared global model from data distributed on edge clients without data transmission.  \fedavg \cite{mcmahan2017communication} is the initial aggregation method, in which parameters of local models trained with data on clients are averaged coordinate-wisely.  Follow-up studies \cite{karimireddy2020scaffold, li2018federated, li2021model, acar2021federated} tackles the client drift mitigation issue \cite{karimireddy2020scaffold} in which local optimums far away from each other when the global model is optimized with different local objectives, and the average of the resultant client updates then move away from the true global optimum. Data-sharing methods including  
As for the aggregation schema, many methods require ideal assumptions such as Lipschitz continuity \cite{zhang2020fedpd,li2018federated,karimireddy2020scaffold,mohri2019agnostic,deng2021distributionally,smith2017federated} and convexity property \cite{mohri2019agnostic,zhang2020fedpd,smith2017federated}. Different from these methods, \textit{model fusion},   to learn a unified model from heterogeneous pre-trained local models,  provides an available approach to FL involving deep neural networks.

\subsection{Ensembling methods}
Ensemble methods \cite{breiman1996bagging,wolpert1992stacked,schapire1999brief, dietterich2000ensemble, breiman2001random} ensemble the outputs of different models to improve the prediction performance. However, this kind of approach requires maintaining these models and thus becomes infeasible with limited computational resources in many applications. In the prior study \cite{yurochkin2019bayesian}, the performance of the ensemble method is viewed as the upper extreme of aggregating when limited to a single communication. However, here in this paper,  we analyze neuron disturbing by combining the uniform ensemble method. 

\subsection{Model Fusion}
\label{sec:model_fusion}
Model fusion methods can be broadly divided into two categories. One category is \textit{knowledge distillation} \cite{hinton2015distilling,bucilua2006model,schmidhuber1992learning}, where the key idea is to employ the knowledge of pre-trained teacher neural networks (local models) to learn a student neural network (global model). In \cite{lin2020ensemble}, the authors propose 
ensemble distillation for model fusion via training the global model through unlabeled data on the outputs of local models.
And in \cite{chen2020fedbe}, the authors sample higher-quality global models and combine them via a Bayesian model. Moreover, a data-free knowledge distillation approach \cite{zhu2021data} is proposed in which the global model learns a generator to assemble local information. Methods based on distillation are generally highly computationally complex and may violate privacy protection because the distillation process in the global model needs either extra proxy datasets or generators. Another category is \textit{parameter matching}, where the key idea is matching the parameters with inherent permutation invariance from different local models before aggregating them together. In \cite{singh2020model}, the authors utilize optimal transport to minimize a transportation cost matrix to align neurons across different neural networks (NNs).  Some work \cite{claici2020model} optimizes the assignments between global and local components under a KL divergence through variational inference. Liu et al. \cite{liu2022deep} formulate the parameter matching as a graph matching problem and solve it with the corresponding method. 
Yurochkin et al. \cite{yurochkin2019statistical} develop a Bayesian nonparametric meta-model to learn shared global structure among local parameters. The meta-model treats the local parameters as noisy realizations of global parameters and formally characterizes the generative process through the Beta-Bernoulli process (BBP) \cite{thibaux2007hierarchical}. This meta-model is successfully extended to different applications \cite{yurochkin2019bayesian, wang2020federated, yurochkin2018scalable, yurochkin2019statistical}. The following algorithms formulate the fusing problem into optimal transport \cite{singh2020model, nguyen2021model} and graph matching \cite{liu2022deep}. However, the above methods rarely investigate the feasibility of model fusion and they are limited to fusing neural networks with identical depth. Some of those algorithms such as OTfusion \cite{singh2020model, nguyen2021model} rely on multiple communication rounds.

\subsection{Data Heterogeneity}
In real-world scenarios, models on varying clients are often trained with heterogeneous datasets, i.e., the data distribution from clients is non-IID. 
There are several categories of non-identical client distributions, including covariate shift, prior probability shift, concept shift, and unbalancedness \cite{kairouz2021advances}. Most previous empirical work on synthetic non-IID datasets \cite{mcmahan2017communication, karimireddy2020scaffold, li2018federated, li2021model, acar2021federated, lin2020ensemble, singh2020model, nguyen2021model, yurochkin2019statistical, nguyen2021model, liu2022deep} have focused on label distribution skew, i.e., prior probability shift, where a non-IID dataset is formed by partitioning an existing IID dataset based on the labels. In this paper, we focus on data heterogeneity resulting from non-IID label distribution.

\section{Preliminaries}
\par Suppose there are $J$ training data sets $\mathcal{D}_j$, $j=1,2,\dotsc, J$ which are sampled from data set $\mathcal{D}=\{(X_i, Y_i)\}, i=1,2,\dotsc, N$ in a non-IID way, where $X_i\in \mathbb{R}^I$ and $Y_i\in \mathbb{R}^C$ is in one-hot coding of classes. The non-IID setting implies that these data distributions of labels from varying data sets are quite different, i.e., heterogeneous.  And the total number of samples is $N:= \sum_{j=1}^J \abs{\mathcal{D}_j}$ where $\abs{\mathcal{D}_j} = N_j$ for $j= 1,2, \dotsc, J$. Besides, we denote $\tilde{\mathcal{D}}_j$ as the corresponding test data set of  $\mathcal{D}_j$, $j=1,2,\dotsc, J$, and the test data set is assumed to be sampled from the same distribution as its corresponding training data set. 
The union of all the test data sets is $\tilde{\mathcal{D}}$, and the total number of sample in this test data set is $\tilde{N}:= \sum_{j=1}^J \abs{\tilde{\mathcal{D}}_j}$,  where $\abs{\tilde{\mathcal{D}}_j} = \tilde{N}_j$ for $j= 1,2, \dotsc, J$.
Without loss of generality, we suppose that $J$ Multilayer Perceptrons with two hidden layers are trained with data sets $\mathcal{D}_j$, $j=1,2,\dotsc, J$, respectively. For the $j$th MLP, let $(W^{(j,0)}\in \mathbb{R}^{I\times L_{j,0}} , b^{(j,0)}\in \mathbb{R}^{L_{j,0}})$,  $(W^{(j,1)}\in \mathbb{R}^{L_{j,0}\times L_{j,1}}, b^{(j,1)}\in \mathbb{R}^{L_{j,1}})$, $(W^{(j,2)}\in \mathbb{R}^{L_{j,1} \times C}, b^{(j,2)}\in \mathbb{R}^{C})$ be the weight and bias pairs of the two hidden layers and softmax layer, respectively. Thus, the $J$th MLP is
\begin{equation*}
    F_j(X) = \mathrm{softmax}\Bigl(W^{(j,2)}\sigma\big(W^{(j,1)}\sigma(W^{(j,0)}X + b^{(j,0)}) + b^{(j,1)}\bigr) + b^{(j,2)}\Bigr), j=1,2, \dotsc, J,
\end{equation*}
where $\sigma$ is the nonlinear activation functions such as ReLU \cite{nair2010rectified}.  For simplicity,  in our theoretical analysis, the bias is neglected by default in this paper via the following augmentation:
\begin{equation*}
    \tilde{W}^{(j,k)} \tilde{a}^{(j, k)} = \begin{pmatrix} W^{(j,k)} & b^{(j,k)}
    \end{pmatrix}
    \begin{pmatrix} a^{(j, k)} \\
    1
    \end{pmatrix}
    = W^{(j,k)} a^{(j, k)} + b^{(j,k)},
\end{equation*}
where $a^{(j, k)}$ is the input of the $k$th layer in the $j$th MLP. Therefore, we only consider MLPs without bias, that is
\begin{equation}
    F_j(X) = \mathrm{softmax}\Bigl(W^{(j,2)}\sigma\big(W^{(j,1)}\sigma(W^{(j,0)}X) \bigr) \Bigr), j=1,2, \dotsc, J,
\end{equation}

The task of fusing neural networks is to learn a global neural network with weights $\theta^{(0)}\in \mathbb{R}^{I\times L_{0}}$,  $\theta^{(1)}\in \mathbb{R}^{L_{0}\times L_{1}}$, $\theta^{(2)}\in \mathbb{R}^{L_{1} \times C}$.


\section{Heterogeneous Neuron Disturbing}
In previous model fusion methods such as \PFNM \cite{yurochkin2019bayesian} and OTfusion \cite{singh2020model}, the authors implicitly set an assumption that, for each layer, 
if the probability measurement of neurons in local models are well approximated by that of the fused global model in a specific sampling manner, then the performance of those local models is also maintained. However, whether this assumption holds for neural networks has not been investigated. In the following, we give a simple counter-example of neuron approximation assumption, and reveal that
neurons from heterogeneous models disturb each other with varying severity which depends on the data heterogeneity. And the neural network architecture is supposed to be adjusted to make the above assumption hold. 

\subsection{Neuron Disturbing from Optimization Unbalance}
\label{sec_nd_opt}
\par
\begin{figure*}[!t]
\centering
\includegraphics[width=0.95\textwidth]{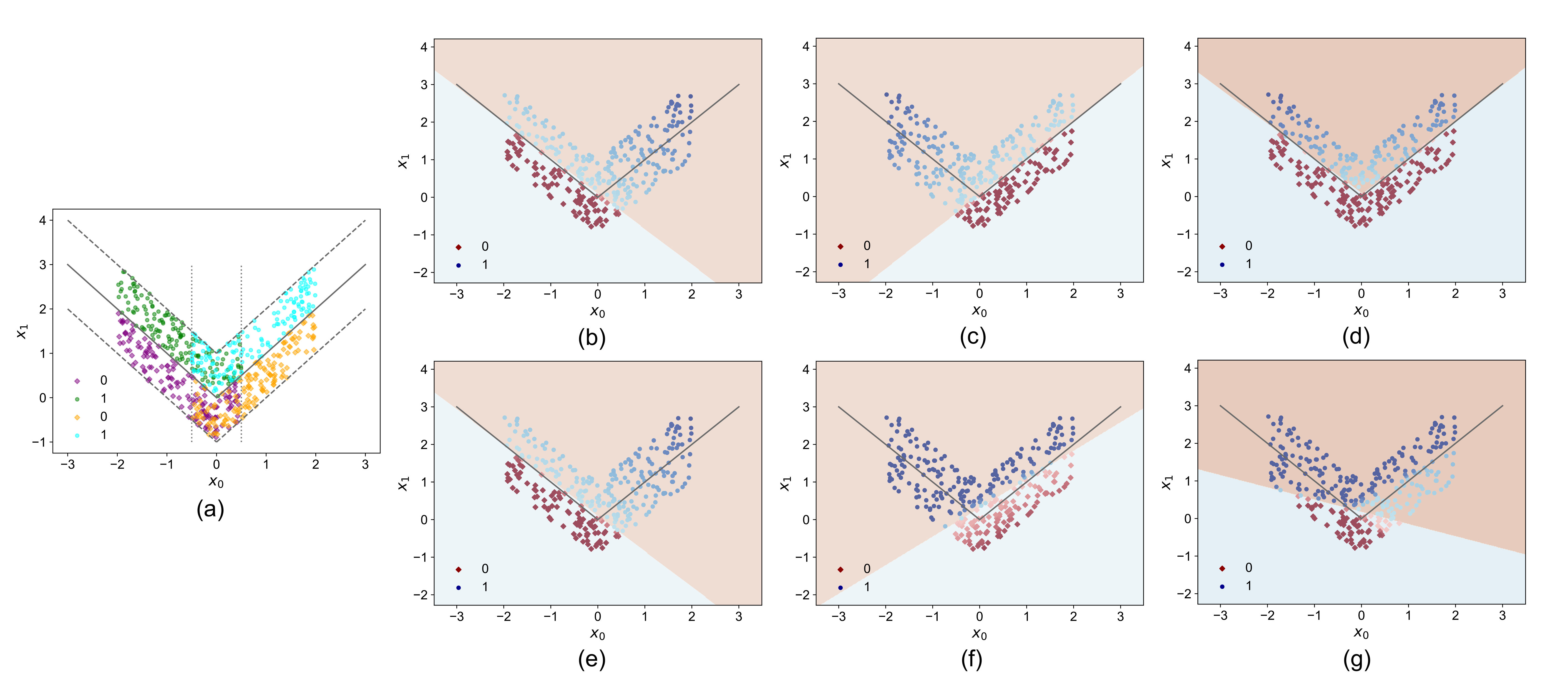}
\caption{The simple example of fusing one-neuron neural networks. (a). The data partition for two local models. The dotted gray lines are $x_1=-0.5$ and $x_1=0.5$, and the dashed dark gray lines are upper and lower bound of the data sampling field.  (b)-(d). A success case of fusing models. (b) and (c) illustrate results of local models on all the test data, and (d) presents the performance of the successfully fused global model. (e)-(g). A fail case of fusing neural networks. Corresponding to (b)-(f) show results of local models on all the test data, and (d) presents the performance of the fail fused global model. The training data, hyper-parameters and neural networks of the two cases are identical except that the random seeds of training are different. }
\label{fig1_example}
\end{figure*}

 We firstly consider a simple binary classification problem on 2D simulation data set. As shown in Figure~\ref{fig1_example}(a), the training and test data samples are randomly sampled from the selected region. The region is bounded by $x_2=-x_1+1,x_1\in [-2, 0)$, $x_2=x_1+1,x_1\in [0, 2]$, $x_2=-x_1-1,x_1\in [-2, 0)$ and $x_2=x_1-1,x_1\in [0, 2]$. We define a decision boundary $C_l$ for data labelling, i.e., $x_2=-x_1,x_1\in [-2, 0)$ and $x_2=x_1,x_1\in [0, 2]$. The samples $(x_1,x_2)$ above $C_l$ are labeled $1$ else labeled $0$. For the left data set, we set a right boundary $x_1 = 0.5$, and sampling $300$ training samples and $150$ test samples, i.e., green and purple points in Figure~\ref{fig1_example}(a). Similarly, for the right data set, we set a left boundary $x_1 = -0.5$, and sampling $300$ training samples and $150$ test samples, i.e., cyan and orange points in Figure~\ref{fig1_example}(a). Therefore, after merging the left and right data sets, The training data set size is $600$ and test data set size is $300$.  

\par After data generation, we train the following neural network with the left and right data sets, respectively. The loss is cross-entropy, and the optimizer is Adam with learning rate equals to $0.5$ and epochs $600$.  
\begin{align*}
\begin{bmatrix}
p^{(j)}_1 \\  p^{(j)}_2
\end{bmatrix} = \mathrm{softmax} \Biggl( \begin{bmatrix}
y^{(j)}_1 \\  y^{(j)}_2
\end{bmatrix} \Biggl), \quad 
\begin{bmatrix}
y^{(j)}_1 \\  y^{(j)}_2
\end{bmatrix} =
\begin{bmatrix}
w^{(j2)}_1 \\
w^{(j2)}_2 
\end{bmatrix} \sigma\Bigl(\begin{bmatrix}
w^{(j1)}_1 &  w^{(j1)}_2
\end{bmatrix} \begin{bmatrix}
x_1 \\ x_2 
\end{bmatrix} + \begin{bmatrix}
b^{j1}
\end{bmatrix} \Bigr), \quad j=1,2.
\end{align*}
Then, we fuse the two trained neural networks into the following global model directly by concatenating the weights and bias. 
\begin{align*}
\begin{bmatrix}
p_1 \\  p_2
\end{bmatrix} = \mathrm{softmax} \Bigl( \begin{bmatrix}
y_1 \\  y_2
\end{bmatrix} \Bigl), \quad 
\begin{bmatrix}
y_1 \\  y_2
\end{bmatrix} = \begin{bmatrix}
w^{(12)}_1 & w^{(22)}_1\\
w^{(12)}_2 & w^{(22)}_2
\end{bmatrix} \sigma\Bigl(\begin{bmatrix}
w^{(11)}_1 &  w^{(11)}_2\\
w^{(21)}_1 &  w^{(21)}_2
\end{bmatrix} \begin{bmatrix}
x_1 \\ x_2 
\end{bmatrix} + \begin{bmatrix}
b^{11} \\ b^{21}
\end{bmatrix} \Bigr).
\end{align*}
The above fusing process essentially imply the assumption that if the probability measurement of neurons in local models is well approximated by the fused global model in specific sampling manner, then the performance of those local models is also maintained. However, in our experiments as shown in Figure~\ref{fig1_example}, we test those two trained local models and their fused model on the merged test data set.  We just repeat the process under different random seeds, and for most of the time, the fused global fails to maintain the predict ability of local models though the local models return almost identical accuracy between the success case (left model:  $ 80.59\%$; right model:$78.95\%$; global model: ${\bm{98.36}}\%$) and fail case (left model:$ 80.59\%$; right model:$ 80.26\%$; global model:${\bm{73.36}}\%$). We also test different activation functions (e.g., ReLU and Leaky ReLU), training hyper-parameters (e.g., set different learning rate and epoch number for thw two models), data partition (e.g., change the dotted gray line to $x_1=-0.25$ and $x_1 = 0.75$) and data amount distribution (e.g., set the size of one dataset much larger than another one), the phenomenon remains. 

\par Furthermore, as seen in Figure~\ref{fig1_example}(c) and Figure~\ref{fig1_example}(f), the two right models have different softmax probability on the real labels, indicated by the darkness of the point color. In the fail case,  the right model shows significantly divergent patterns of softmax probability compared to the left model, i.e., has lower softmax values for label $1$ and higher softmax values for label $0$ samples. While in the success case, the left and right model output close softmax values for label $1$ and label $0$ samples. Actually, the output of the global model is the dot addition of the local models, i.e.,
\begin{equation}
    \begin{bmatrix}
y_1 \\  y_2
\end{bmatrix} = \begin{bmatrix}
y^{(1)}_1 +  y^{(2)}_1\\
y^{(1)}_2 + y^{(2)}_2
\end{bmatrix}.
\end{equation}
Hence, for two neural networks, the output depends on how their outputs diverge. Due to the data heterogeneity, the quality of the solutions of the two local models may differs. For cross-entropy loss as below, as the optimizing goes on, for the true lable $c$, the corresponding softmax probability $\hat{p}_{ic}$ converges to $1$, and for the other labels,  $\hat{p}_{ic'}$($c'\neq c$) converges to $0$. As a consequence, for samples those only display in left or right data set, the scale of the outputs from two neural networks is quite different. For example, for a sample of label $0$ in Figure~\ref{fig1_example}(e) which is classified as label $1$ in Figure~\ref{fig1_example}(g), if $(y^{(1)}_1, y^{(1)}_2)=(4, -5)$ and $(y^{(2)}_1, y^{(2)}_2)=(-2, 2)$, the output $(y_1, y_2)=(2, -3)$. Thus in the fail case, prediction results of the global model are dominated by the first local model. 
\begin{equation}
\operatorname{Cross-Entropy(\{Y_i\}, \{\hat{Y}_i\})} = \frac{1}{N}\sum_{i=1}^N \sum_{c=1}^C \left( -Y_{ic} \log(\hat{p}_{ic}) \right),
\end{equation} 
where $\hat{p}_{ic}$ is the $c$th element of the estimated softmax probability vector, and $Y_{ic}$ is the $c$th element of $Y_i$. 

\subsection{A Bayesian View of Neuron Disturbing}
We now give a basic explanation of neuron disturbing from the Bayesian viewpoint.  Assume $\bmf{\theta}_1, \bmf{\theta}_2, \dotsc, \bmf{\theta}_J $ are parameters of $J$ heterogeneous local models, respectively. The inference process of those parameters based on datasets $\mathcal{D}_j$, $j=1,2,\dotsc, J$ can be represented by posterior $P_j(\bmf{\theta}_j \vert \mathcal{D}_j)$. And the parameters of the fused global model is denoted as $\bmf{\theta}$. 
The model fusion process is formulated as
\begin{align*}
    P(\bmf{\theta} \vert \{\mathcal{D}_1, \mathcal{D}_2, \dotsc, \mathcal{D}_J \}) = P(\bmf{\theta} \vert \{\bmf{\theta}_1, \bmf{\theta}_2, \dotsc, \bmf{\theta}_J \}) \, P(\{\bmf{\theta}_1, \bmf{\theta}_2, \dotsc, \bmf{\theta}_J \vert \{\mathcal{D}_1, \mathcal{D}_2, \dotsc, \mathcal{D}_J \}).
\end{align*}
The forward propagation of $\mathcal{D}_j$  in neural network $F_j$ is 
\begin{equation}
    P(\mathcal{D}_j \vert \bmf{\theta}_j) = P\bigl(F_j(X \vert \bmf{\theta}_j) = Y \big\vert (X, Y) \in \mathcal{D}_j \bigr), j = 1,2, \dotsc, J.
\end{equation}
Hence, the likelihood of all the local datasets and local models is
\begin{align}
\label{eq_local_likeli}
    P(\{ \mathcal{D}_1, \mathcal{D}_2, \dotsc, \mathcal{D}_J \} \big\vert \{\bmf{\theta}_1, \bmf{\theta}_2, \dotsc,  \bmf{\theta}_J\}) 
   =  \prod_{j=1}^J P( \mathcal{D}_j \big\vert \bmf{\theta}_j) 
   =  \prod_{j=1}^J \frac{P\bigl( \bmf{\theta}_j\big\vert \mathcal{D}_j \bigr) P(\mathcal{D}_j)}{P(\bmf{\theta}_j)}
    = \frac{\prod_{j=1}^J P(\bmf{\theta_j} \vert \mathcal{D}_j)  \prod_{j=1}^J P( \mathcal{D}_j)}{\prod_{j=1}^J P(\bmf{\theta_j} )}.
\end{align}
On the other side, the likelihood of the global model and the datasets is
\begin{align}
\label{eq_global_likeli}
    P(\{ \mathcal{D}_1, \mathcal{D}_2, \dotsc, \mathcal{D}_J \} \big\vert \bmf{\theta})
    = \frac{ P(\bmf{\theta} \vert \{ \mathcal{D}_1, \mathcal{D}_2, \dotsc, \mathcal{D}_J \})   P( \{ \mathcal{D}_1, \mathcal{D}_2, \dotsc, \mathcal{D}_J \})}{ P(\bmf{\theta} )}.
\end{align}

Without loss of generality, the following equation is assumed to hold, since here we investigate how concatenated neurons disturb each other. 
\begin{align*}
    P(\bmf{\theta}) = \prod_{j=1}^J P(\bmf{\theta_j} ). 
\end{align*}

However, even with this assumption, equation~(\ref{eq_global_likeli}) generally does not equal to equation~(\ref{eq_local_likeli}), because for most of the time
\begin{align*}
    P( \{ \mathcal{D}_1, \mathcal{D}_2, \dotsc, \mathcal{D}_J \}) \neq \prod_{j=1}^J P( \mathcal{D}_j),
\end{align*}
and 
\begin{align*}
    P(\bmf{\theta} \vert \{ \mathcal{D}_1, \mathcal{D}_2, \dotsc, \mathcal{D}_J \}) \neq \prod_{j=1}^J P(\bmf{\theta_j} \vert \mathcal{D}_j).
\end{align*}
These inequalities generally holds since local datasets $\mathcal{D}_1, \mathcal{D}_2, \dotsc, \mathcal{D}_J$ are not Independent and identically distributed. 


\subsection{Absolute Confidence of Neural Networks}
\par Neuron disturbing reveals the drawback of those methods that directly fusing neurons of heterogeneous local models. Although there is disturbing among neurons from different local models trained with varying datasets, we found another phenomenon which inspires us to avoid neuron disturbing. 
\par
\begin{figure*}[!t]
\centering
\includegraphics[width=1\textwidth]{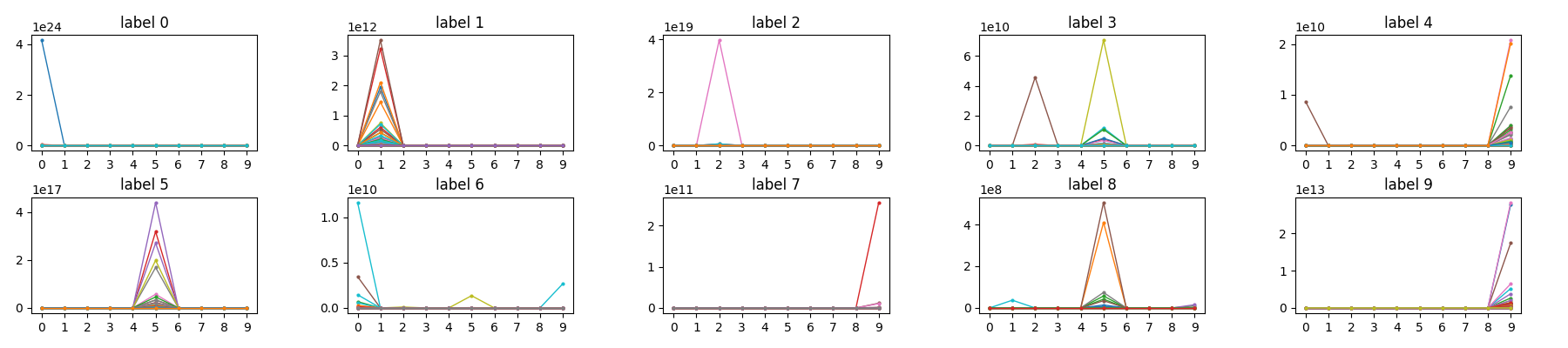}
\caption{Exponential of output on all the test data of MNIST before softmax of 3-layer MLP trained with label 0,1,2,5 and 9 of  MNIST training data. }
\label{fig2_example}
\end{figure*}

 We firstly train a 3-layer MLP with label 0,1,2,5 and 9 data in MNIST training data. Then, the trained model is tested on test data of MNIST spanning all 10 labels. Denote the model as $F_j(X)$, where $X$ is the input and $j$ is the index of local model. Besides, we denote the output of layer before softmax as $f_j(X) = \bigl(f^{(1)}_j(X), f^{(2)}_j(X), \dotsc, f^{(C)}_j(X) \bigr)^{\mathrm{T}}$. For each test sample, we compute the exponential of output $f_j(X)$, i.e., $\exp{f_j(X)} = \bigl(\exp{f^{(1)}_j(X)}, \exp{f^{(2)}_j(X)}, \dotsc, \exp{f^{(C)}_j(X) \bigr)^{\mathrm{T}}}$. Since the index of the maximum value of $f_j(X)$ or $\exp{f_j(X)}$ determine the predicted label of the input, we present the maximum value of $\max_{c=1}^C\exp{f^{(c)}_j(X)}$ for all the test samples in figure~\ref{fig2_example}. We name it as absolute confidence of $F_j$ on sample $X$. As illustrated in figure~\ref{fig2_example}, the scales of absolute confidence of most samples of label $0$,$1$,$2$,$5$ and $9$ ($10^{12}$ to $10^{24}$) are significantly higher than that of the other samples ($10^{8}$ to $10^{11}$). This phenomenon implies that, for a sample with specific label, it is possible to select a local model trained with data with this label. Therefore, in this way, the selected model performs better than the other models. In summary, local model's output with highest absolute confidence on $X$ indicates the superior classification ability of this local model among all the models. We do experiments on MNIST and CIFAR10 with different neural networks including MLPs and CNNs, the phenomenon remains with diverse hyper-parameters settings. 


\section{Adaptive Model Selection}
\subsection{General procedure}
From above analysis, a natural way of fusing heterogeneous neural networks trained with silo data sets is to select a local model depending on the input $X_i$. The local model trained with more samples close to $X_i$ has higher activation values in each layer, especially the last layer before the softmax operation.  Hence, we achieve adaptive model selection via comparing the maximum output values of each local models, i.e., the absolute confidence values, and selecting the local model with the highest maximum output value. Formally, for the disturbing matrix
\begin{equation}
    M^{(JC)}(x) = \begin{bmatrix}
    y^{(1)}_1 &  y^{(2)}_1 & \ldots & y^{(j)}_1 \\
y^{(1)}_2 & y^{(2)}_2 & \ldots & y^{(j)}_2 \\
\vdots & \vdots & \ldots & \vdots \\
y^{(1)}_C & y^{(2)}_C & \ldots & y^{(j)}_C
    \end{bmatrix} = \begin{bmatrix}
    f^{(1)}_1(x) &  f^{(2)}_1(x) & \ldots & f^{(J)}_1(x) \\
f^{(1)}_2(x) & f^{(2)}_2(x) & \ldots & f^{(J)}_2(x) \\
\vdots & \vdots & \ldots & \vdots \\
f^{(1)}_C(x) & f^{(2)}_C(x) & \ldots & f^{(J)}_C(x)
    \end{bmatrix} = \begin{bmatrix}
    f^{(1)}(x) &  f^{(2)}(x) & \ldots &  f^{(J)}(x) 
    \end{bmatrix}.
\end{equation}
The maximum activation value vector of each local model is then 
\begin{equation}
    \begin{bmatrix}
    f^{(1)}_*(x) &  f^{(2)}_*(x) & \ldots &  f^{(J)}_*(x) 
    \end{bmatrix} = \begin{bmatrix}
    \max_{c_1} f^{(1)}_{c_1}(x) &  \max_{c_2} f^{(2)}_{c_2}(x) & \ldots &  \max_{c_J} f^{(J)}_{c_J}(x)
    \end{bmatrix}
\end{equation}
Then, the local model index $j^*_{i}$ for sample $X_i$ is chosen via another maximizing process on the maximum activation value vector, i.e., 
\begin{equation}
    j^*_{i} = \arg\max_{1 \leq j \leq J} f^{(j)}_*(X_i),
\end{equation}

\subsection{Algorithm}
The above procedure leads to two experimental algorithms. 
As shown in algorithm~\ref{ams_same}, for models with the same architecture (the same depth), the weights of the $l$th layer in local models are put into a global weight matrix $W^{(l)}$. This operation keeps neural networks from disturbing each other until the last layers. However, this operation here is for clear explanation of neuron disturbing and not necessary. As for the last layer, \AMS select the local model corresponding to input $X$ via comparing the maximum activation value of all the models before softmax operation. This is also called \AMStop in the following part. 

\par For comparison, the maximum one operation is replaced by summation of the maximum $k$ activation vectors from those local models. When $k$ equals to the number of models, the algorithm is equivalent to the general fusion process in previous researches, except the fusion only happens in the last layer. The other parts of those local models are independent. We call this algorithm \AMSfull, and it provides a fused model where neuros disturbing only happens in the last layer.  Algorithm~\ref{ams_cross} is an extended version for ensembling nets with different architectures. 

\renewcommand{\algorithmicrequire}{ \textbf{Input:}} 
\renewcommand{\algorithmicensure}{ \textbf{Output:}} 
\begin{algorithm}
\caption{Adaptive model selection for models with the same architecture}
\label{ams_same}
\begin{algorithmic}[1]
    \REQUIRE ~~\\
    Trained local models $\{F^{(j)}\}$ from $S$ clients input $X$
    \ENSURE ~~\\
    Label of input $Y$;

    \STATE   The first layer weight of the global model $W^{(0)} = \bigl[{W^{(10)}}^{\rm{T}}, {W^{(20)}}^{\rm{T}}, \dotsc, {W^{(J0)}}^{\rm{T}} \bigr]^{\rm{T}}$
     \FOR{$l$ = 2, $\dotsc$, L-1}
         \STATE  The $l$th layer weight of the global model $W^{(l)} = \mathrm{Diag} \bigl[{W^{(1l)}}, {W^{(2l)}}, \dotsc, {W^{(Jl)}} \bigr]$
    \ENDFOR  
    \STATE  $F^{(L-1)}(X) = \Bigl(W^{(L-1)} \dotsm \sigma\big(W^{(1)}\sigma(W^{(0)}X) \bigr) \Bigr)$
     \STATE The output layer weight of the global model $W^{(L)} = \bigl[{W^{(1L)}}, {W^{(2L)}}, \dotsc, {W^{(JL)}} \bigr]$
     \STATE $j^* = \arg\max_{1 \leq j \leq J} \begin{bmatrix}
    \max_{c_1}^C\bigl({W^{(1L)}}F^{(L-1)}(X)\bigr)_{c_1} &  \max_{c_2}^C \bigl({W^{(2L)}}F^{(L-1)}(X)\bigr)_{c_2} & \ldots &  \max_{c_J}^C \bigl({W^{(JL)}}F^{(L-1)}(X)\bigr)_{c_J}
    \end{bmatrix} $
    \STATE The output is $Y = F(X) = \mathrm{softmax}\bigl(F^{(L)}_{:,j^*}(X) \bigr)$
\end{algorithmic}
\end{algorithm}

\renewcommand{\algorithmicrequire}{ \textbf{Input:}} 
\renewcommand{\algorithmicensure}{ \textbf{Output:}} 
\begin{algorithm}
\caption{Adaptive model selection for models with different architectures}
\label{ams_cross}
\begin{algorithmic}[1]
    \REQUIRE ~~\\
    Trained local models $\{F^{(j)}\}$ from $S$ clients;input $X$
    \ENSURE ~~\\
    Label of input $Y$
    \STATE Extract local model $\{f^{(j)}\}$ without softmax operation from $\{F^{(j)}\}$, $j=1,2,\dotsc, J$
    \STATE $j^* = \arg\max_{1 \leq j \leq J} \begin{bmatrix}
    \max_{c_1}^Cf^{(1)}_{c_1}(X) &  \max_{c_2}^C f^{(2)}_{c_2}(X) & \ldots &  \max_{c_J}^C f^{(J)}_{c_J}(X)
    \end{bmatrix} $
    \STATE The output of the global model is $F = \mathrm{softmax}\bigl(f^{(j^*)}(X)\bigr)$
\end{algorithmic}
\end{algorithm}

\subsection{Computational complexity reduction}
The method proposed in this paper, \AMS, is kind of an intermediate method between ensemble method and model fusion. The execution of \AMS require saving all the local models and all the predictions of those models on the input. The high computational and store cost is unacceptable in real-world scenario. However, it is straightforward to reduce the computational complexity of \AMS, and extend it to a typical model fusion method via the following techniques. 
\begin{itemize}
    \item Zero-shot knowledge distillation \cite{nayak2019zero}. Unlike those distillation methods introduced in section~\ref{sec:model_fusion}, some distillation methods do not require additional datasets and this facilitates distilling \AMS into a light-weight model.  
    \item Client selection. Despite of the heterogeneity among clients, it is common that a portion of clients share similar training data, especially when the clients amount is huge. This motivates recent researches \cite{nishio2019client, cho2020client, cho2022towards} to select clients before model aggregation. Client selection before executing \AMS is conductive to decrease computational complexity.
    \item Adaptive model selection on the first layer. Our experiments and the design of \AMS are mainly for  confirmation of neuron disturbing. However, the maximum operation on absolute confidence of neural networks is still effective on the layers besides of the last layer, although the fusing performance decays. We leave it for future work.
\end{itemize}

\section{Experiments}
This section presents empirical validations of our analysis, including the existence of neuron disturbing and the practicability of the absolute confidence of neural networks. We test the performance of classical methods of different types on image classification task. The experiments involve settings of varying data heterogeneity and client amount, distinctive types of neural networks including MLPs and CNNs, and different net architectures. 

\subsection{Setup}
\textbf{Datasets and models}. We evaluate our algorithm on MNIST \cite{lecun2010mnist} and CIFAR 10 \cite{krizhevsky2009learning}, two standard image classification datasets and each contains ten classes on handwriting digits and objects in real life, respectively. For MNIST, we apply a MLP model with varying number of hidden layers; for CIFAR 10, we apply a ConvNet with 3 convolutional and 2 fully-connected layers. 

\textbf{Partition strategies of client data}. Here we consider two heterogeneous partition strategies, \textit{hetero-label} and \textit{hetero-dir},  to simulate federated learning scenarios where the number of data points and class proportions in each client is unbalanced. In \textit{hetero-label} partition, each client is randomly assigned all the samples with $3$ to $6$ labels for MNIST and CIFAR10. 
In \textit{hetero-dir} partition, for CIFAR10 and MNIST, we follow prior works \cite{yurochkin2018scalable} which apply $K$-dimensional Dirichlet distribution $Dir(\alpha)$ to create non-iid data, in which a smaller $\alpha$ indicates higher data
heterogeneity. Specifically, for dataset with class number $K$, we sample the proportion of the instances of class $k$ to client $s$, $p_{k,s}$, via $p_{k,s} \sim Dir_k(\alpha)$, where $K = 10$ and $\alpha = 0.5$ by default. 
In each dataset, we execute 5 trials to obtain mean and standard deviations of the performance. For fairness, all the algorithms are executed on the same data for each setting.  

\textbf{Baselines}. We compare our method with original \PFNM \cite{yurochkin2019bayesian}, FedAvg \cite{mcmahan2017communication}, and uniform ensemble method. The hyperparameters of \PFNM are optimally adjusted.  Here FedAvg is operated in local neural networks trained with the same random initialization as proposed by \cite{mcmahan2017communication}.

\textbf{Training setup} Our method and baselines operate on the collection of neural network weights from all partitioned batches. We use PyTorch \cite{paszke2017automatic} to implement these networks and train them by the Adam optimizer \cite{kingma2014adam}. All hyperparameter settings are summarized in table~\ref{tabel_run_params}.

\begin{table}[htb]
\caption{Hyperparameter settings for training neural networks}
 \label{tabel_run_params}
\begin{center}
 \begin{tabular}{ccccccc}
  \toprule
 Dataset & Model  & Optimizer & Learning rate (decay, period)  & Batch size & Epochs & Regularization \\
  \midrule
  MNIST & MLP & Adam & 0.001 (0.8, 2) & 64 & 40 & $L_1$ ($10^{-7}$) \\
  \midrule
  CIFAR10  & ConvNet & Adam & 0.01 (0.8, 3) & 128 & 50 & $L_1$ ($10^{-7}$) \\
 \bottomrule
 \end{tabular}
\end{center}
\end{table}

\subsection{Experiments Results}
\par We evaluate \AMS and other model fusion or ensemble methods and the impact of neural disturbing via three groups of experiments: (1) heterogeneous MLPs and CNNs with the same depth, (2) heterogeneous MLPs with the varying depth, and (3) fusing MLPs and CNNs with varying severity of data heterogeneity. The code is available at \url{https://github.com/Codsir/ams}.

\noindent \textbf{Fusing heterogeneous models with the same depth} As shown in table~\ref{table_hetero_label} and table~\ref{table_hetero_dir}, for all the settings, the accuracy of global model fused via \AMStop is significantly higher than that of \AMSfull. Since the only difference between \AMStop and \AMSfull is that \AMSfull does not exclude mutual correlations of neurons in the last layer of the nets,  this demonstrate that neuron disturbing exists and has a obvious impact on model fusion. Even \AMSfull does not exclude neuron disturbing before the output, the performance of \AMSfull is already competitive with \fedavg and \PFNM. The reflects that it is possible the performance of those methods is upper bounded by neural disturbing. 

\par Another observation is that performance of ensemble method is not stable under severe heterogeneous data partitions, e.g., it perform significantly worse (about $10\%$ lower for accuracy) than \AMStop in \textit{hetero-label} partition. This indicates that ensemble no longer works in certain scenarios where data is heterogeneous. When viewed as a special kind of ensemble method, \AMS adapts fusion operation on absolute confidence of nets instead of activated outputs. Thus, absolute confidence of neural networks is demonstrated to be informative for model fusion. Besides, compare to the other methods, \AMStop and the ensemble method are robust to client number and net depth, although ensemble method is relatively unstable (accuracy standard deviation is significantly higher for \textit{hetero-dir} partition). This is consistent with above results where ensemble method is robust to data heterogeneity.  Overall, As a mediate method between model fusion and ensemble method,  \AMS performs better than the other methods. 
\begin{table*}[htb]
\caption{Performance overview under hetero-label data partition }
 \label{table_hetero_label}
 \begin{center}
 \setlength\tabcolsep{2pt}
     \begin{tabular}{cccccccc}
      \hline
       \thead{Datasets \\  (Architectures)}  & $\rm{N}$ & $\rm{L}$ & ensemble  & \fedavg & \PFNM & \AMSfull & \AMStop \\
      \hline
        ~  & 5 & 1 & \textit{62.12 $\pm$ 11.89}  & 46.89 $\pm$ 8.08   & \textbf{86.81 $\pm$ 6.05} & 52.82 $\pm$ 9.27 & 81.60 $\pm$ 4.73  \\
       ~ & 10 & 1 & \textit{70.37 $\pm$ 12.12}    & 56.37 $\pm$ 7.31 & \textbf{83.45 $\pm$ 7.50} &59.63 $\pm$ 8.24  & 82.08 $\pm$ 2.95  \\
       MNIST &  20 & 1 & \textit{80.75 $\pm$ 6.74}   & 69.51 $\pm$ 1.52 & 84.00 $\pm$ 3.19 & 75.23 $\pm$ 2.30 & \textbf{87.26 $\pm$ 1.89} \\
       (MLPs) & 30 & 1 & \textit{80.22 $\pm$ 11.20}    & 68.36 $\pm$ 1.35  & \textbf{85.41 $\pm$ 4.54} & 81.09 $\pm$ 2.78 & 82.95 $\pm$ 0.90  \\
       ~ & 10 & 2 &  \textit{69.02 $\pm$ 7.64}   & 61.84 $\pm$ 5.59 &  81.49 $\pm$ 3.86 & 77.85 $\pm$ 11.58 & \textbf{85.82 $\pm$ 3.83}  \\
       ~ &  10 & 3 & \textit{70.57 $\pm$ 11.93}  & 49.67 $\pm$ 8.73 &  80.31 $\pm$ 8.61  & 73.92 $\pm$ 10.90 & \textbf{82.15 $\pm$ 1.47} \\
       ~ &  10 & 4  &  \textit{68.09 $\pm$ 11.13}   & 48.63 $\pm$ 2.53  & 65.91 $\pm$ 7.14 & 70.78 $\pm$ 5.90 & \textbf{79.91 $\pm$ 4.80}  \\
      \hline
       ~ &  5 & 3-2 & \textit{49.01 $\pm$ 5.61}  & 20.77 $\pm$ 7.97 & 33.30 $\pm$ 7.75  & 32.27 $\pm$ 9.12 & \textbf{47.87 $\pm$ 5.44} \\
        CIFAR10  & 10 & 3-2 & \textit{56.14 $\pm$ 7.03}  & 18.05 $\pm$ 2.11 & 30.90 $\pm$ 4.09 & 35.73 $\pm$ 7.69  &  \textbf{56.05 $\pm$ 3.69}  \\
       (ConvNet) & 20 & 3-2 & \textit{61.81 $\pm$ 5.76}   & 15.34 $\pm$ 2.84 & 34.09 $\pm$ 7.04 & 37.79 $\pm$ 8.84 & \textbf{58.22 $\pm$ 0.71} \\
        ~  & 30 & 3-2 & \textit{65.40 $\pm$ 3.36}  & 19.29 $\pm$ 2.64  & 27.84 $\pm$ 4.06  & 46.60 $\pm$ 5.73 & \textbf{50.82 $\pm$ 2.94} \\
     \hline
     \end{tabular}
 \end{center}
\end{table*}

\begin{table*}[htb]
\caption{Performance overview under hetero-dir data partition}
 \label{table_hetero_dir}
 \begin{center}
 \setlength\tabcolsep{2pt}
     \begin{tabular}{cccccccc}
      \hline
       \thead{Datasets \\  (Architectures)}  & $\rm{N}$ & $\rm{L}$ & ensemble & \fedavg & \PFNM & \AMSfull & \AMStop \\
      \hline
        ~  & 5 & 1 & \textit{88.33 $\pm$ 2.54}    & 82.07 $\pm$ 2.74 & 86.96 $\pm$ 5.05 & 85.25 $\pm$ 5.20 & \textbf{92.89 $\pm$ 2.46} \\
       ~ & 10 & 1 & \textit{93.30 $\pm$ 1.28}     & 81.29 $\pm$ 2.27 & 87.49 $\pm$ 2.78 & 82.58 $\pm$ 2.44 & \textbf{90.71 $\pm$ 2.43} \\
       MNIST &  20 & 1 & \textit{88.51 $\pm$ 3.13}   & 82.57 $\pm$ 3.31 & 81.02 $\pm$ 4.56 & 83.78 $\pm$ 3.95 & \textbf{86.66 $\pm$ 1.24}  \\
       (MLPs) & 30 & 1 & \textit{90.57 $\pm$ 1.37}   & 79.49 $\pm$ 2.07 &  81.36 $\pm$ 2.35 & 80.73 $\pm$ 1.83 &  \textbf{84.22 $\pm$ 1.35}  \\
       ~ & 10 & 2 & \textit{93.93 $\pm$ 1.38}  & 82.80 $\pm$ 3.06  &   88.41 $\pm$ 2.42 &  85.72 $\pm$ 3.29 & \textbf{91.12 $\pm$ 1.92}  \\
       ~ &  10 & 3 & \textit{93.93 $\pm$ 1.77}    & 84.01 $\pm$ 3.37  &  89.53 $\pm$ 1.73 & 88.24 $\pm$ 3.20 & \textbf{91.33 $\pm$ 1.23}  \\
       ~ &  10 & 4  & \textit{94.00 $\pm$ 2.03}   & 78.43 $\pm$ 2.81 &  87.23 $\pm$ 2.43 & 87.49 $\pm$ 2.00 & \textbf{89.43 $\pm$ 2.00}  \\
      \hline
       ~ &  5 & 3-2 & \textit{58.86 $\pm$ 4.41}  &20.49 $\pm$ 2.36  &   50.12 $\pm$ 2.28 & 47.08 $\pm$ 3.51 &  \textbf{61.46 $\pm$ 0.89} \\
        CIFAR10 & 10 & 3-2 & \textit{55.74 $\pm$ 1.24}    & 17.89 $\pm$ 3.13 & 49.00 $\pm$ 2.43 & 35.31 $\pm$ 2.46 &  \textbf{54.29 $\pm$ 3.12}  \\
        (ConvNet) & 20 & 3-2 & \textit{47.56 $\pm$ 1.14}  & 16.79 $\pm$ 5.16 & \textbf{44.06 $\pm$ 1.49} & 28.98 $\pm$ 4.34 & 40.08 $\pm$ 3.36  \\
         ~ &  30 & 3-2 & \textit{45.19 $\pm$ 1.10}  & 17.19 $\pm$ 1.71  & \textbf{43.02 $\pm$ 2.72}  & 30.37 $\pm$ 2.17 & 38.89 $\pm$ 2.89  \\
     \hline
     \end{tabular}
 \end{center}
\end{table*}

\noindent \textbf{Fusing heterogeneous models with varying architecture} In previous research, model fusion algorithms are only applicable on neural networks with the same architectures (at least the same depth). However, to a substantial extent, to fuse heterogeneous models with different architectures is in great demand in many scenarios. Due to the special design, \AMS is available for fusing models with diverse architectures only if their outputs share the same dimension. As shown in table~\ref{table_cross_layer},  the performance of \AMStop is just similar to that of fusing models with the same architecture, and \AMStop is applicable to fuse networks with different layers. In addition, the ensemble method is competitive with \AMStop when data severity is relatively low, however, when data heterogeneity is high, the performance of ensemble declines dramatically. It is not hard to interpret since ensemble method does not contain mechanisms for data heterogeneity. Therefore, \AMS is robust to heterogeneity when it is viewed as a special ensemble algorithm. 

\begin{table*}[htb]
\caption{Performance of \AMS for cross-layer fusion}
 \label{table_cross_layer}
 \begin{center}
 \setlength\tabcolsep{2pt}
     \begin{tabular}{cccccccc}
      \hline
       \thead{Data Partition} & Dataset & Net Type & $\rm{N}$ & Net Depth ($\rm{L}$)  & ensemble  & AMS-top1  \\
      \hline
        ~  & ~ & ~ & 5 & $\left[ 1,5 \right]$ & 73.43 $\pm$ 8.15  &   \textbf{81.49 $\pm$ 7.76}  \\
       hetero-label & MNIST & MLPs & 10  & $\left[ 1,5 \right]$  & 57.68 $\pm$ 7.90 &  \textbf{77.32 $\pm$ 4.77}  \\
        ~ &  ~ & ~ &  20 & $\left[ 1,5 \right]$ & 76.40 $\pm$ 4.37  &  \textbf{81.42 $\pm$ 4.84}  \\
      \hline
        ~  & ~ & ~ & 5 & $\left[ 1,5 \right]$ & \textbf{94.98 $\pm$ 1.65}  &   91.68 $\pm$ 2.55 \\
       hetero-dir ($\alpha=0.5$) & MNIST & MLPs & 10  & $\left[ 1,5 \right]$  & \textbf{92.20 $\pm$ 2.81} &   90.90 $\pm$ 1.22 \\
        ~ &  ~ & ~ &  20 & $\left[ 1,5 \right]$ & \textbf{92.75 $\pm$ 1.76} &    88.21 $\pm$ 1.76 \\
     \hline
         ~  & ~ & ~ & 5 & $\left[ 1,5 \right]$ & 45.45 $\pm$ 3.84   & \textbf{61.90 $\pm$ 4.93}   \\
       hetero-dir ($\alpha=0.01$) & MNIST & MLPs & 10  & $\left[ 1,5 \right]$  & \textbf{37.14 $\pm$ 6.50} &  29.26 $\pm$ 12.21  \\
        ~ &  ~ & ~ &  20 & $\left[ 1,5 \right]$ &  25.81 $\pm$ 7.47 &  \textbf{34.10 $\pm$ 2.57} \\     
     \hline
     \end{tabular}
 \end{center}
\end{table*}

\begin{figure*}[!t]
\centering
\includegraphics[width=1\textwidth]{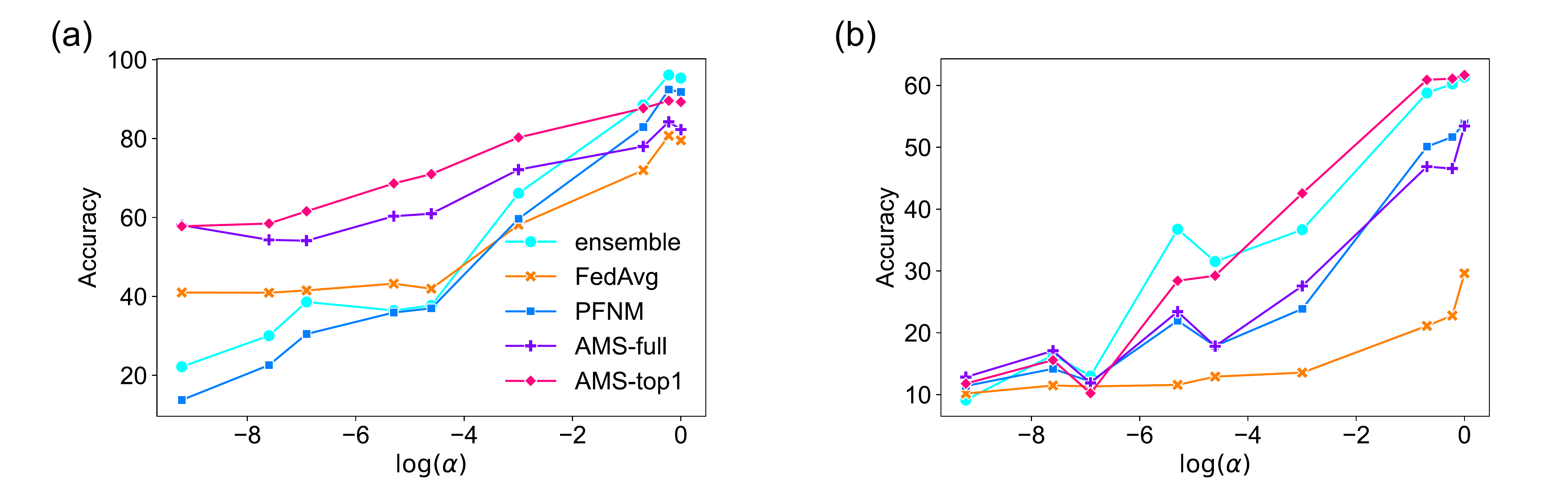}
\caption{Performance comparison of model fusion and ensemble methods on MNIST (a) and CIFAR10 (b) with varying heterogeneity ($\alpha\in [5^{-4}, 1]$) of data partition. }
\vspace{-0.4cm}
\label{fig_hetero_alpha}
\end{figure*}

\noindent \textbf{Heterogeneity robustness} We here test how all the model fusion and ensemble algorithms performs when the severity of data heterogeneity is increased. We run each method on five local MLPs (MNIST) and CNN (CIFAR10) models trained with \textit{hetero-dir} data partition with varying $\alpha$. As shown in figure~\ref{fig_hetero_alpha},  \AMSfull and \AMStop achieves significantly higher accuracy than \fedavg, ensemble method and \PFNM when $\alpha$ is smaller than $0.5$. For MLPs trained with MNIST, \AMStop and \AMSfull maintain a accuracy of close to $60\%$ while accuracy of ensemble method and \PFNM decays to $20\%$ when $\alpha=5^{-4}$. FedAvg maintains a stable accuracy close to $40\%$ when $\alpha$ is smaller than $0.01$.  In contrast, in the figure~\ref{fig_hetero_alpha}(b), for CNNs trained with CIFAR10, the robustness performance rank of all the methods hold similarly while move towards right.  We explain this phenomenon as the resultant of increment of task difficulty. 
 
 \par Considering that the \AMSfull and \AMStop exclude neuron disturbing before the output layer. The above facts show that it is necessary to consider the impact of neuron disturbing in overcoming data heterogeneity for model fusion or ensemble methods. Performance of \fedavg and \PFNM decline rapidly when the severity of data heterogeneity reaches some point.  Considering preceding model fusion method are almost all only tested on \textit{hetero-dir} data partition where $\alpha \geq 0.5$, we doubt the reliability of experiments in previous researches. Furthermore, it is sensible to set robustness to data heterogeneity as a standard evaluation index when newly developed model fusion algorithms are tested.

\section{Conclusion}
\label{conclusion}
 In this paper, we reveal the phenomenon of neuron disturbing and give detailed explanations from a Bayesian viewpoint combining the data heterogeneity among clients and the property of neural networks. In addition, based on absolute confidence of neural networks, we developed a experimental method for verification of neural disturbing. There are still several concerns left for future works. The first one is, how to measure neural disturbing and characterize the dynamics of it along the change of data heterogeneity. The second, what is the sufficient condition of effectiveness of \AMS method, e.g., the loss function.  The third, if there is disturbing among learned parameters in statistical models trained with non-IID data. 


\bibliographystyle{unsrt}  
\bibliography{ams}

\begin{thebibliography}{10}

\bibitem{mcmahan2017communication}
Brendan McMahan, Eider Moore, Daniel Ramage, Seth Hampson, and Blaise~Aguera
  y~Arcas.
\newblock Communication-efficient learning of deep networks from decentralized
  data.
\newblock In {\em Artificial Intelligence and Statistics}, pages 1273--1282,
  2017.

\bibitem{li2020federated}
Tian Li, Anit~Kumar Sahu, Ameet Talwalkar, and Virginia Smith.
\newblock Federated learning: Challenges, methods, and future directions.
\newblock {\em IEEE Signal Processing Magazine}, 37(3):50--60, 2020.

\bibitem{li2018federated}
Tian Li, Anit~Kumar Sahu, Manzil Zaheer, Maziar Sanjabi, Ameet Talwalkar, and
  Virginia Smith.
\newblock Federated optimization in heterogeneous networks.
\newblock {\em arXiv preprint arXiv:1812.06127}, 2018.

\bibitem{utans1996weight}
Joachim Utans.
\newblock Weight averaging for neural networks and local resampling schemes.
\newblock In {\em Proc. AAAI-96 Workshop on Integrating Multiple Learned
  Models. AAAI Press}, pages 133--138. Citeseer, 1996.

\bibitem{yurochkin2019bayesian}
Mikhail Yurochkin, Mayank Agarwal, Soumya Ghosh, Kristjan Greenewald, Nghia
  Hoang, and Yasaman Khazaeni.
\newblock Bayesian nonparametric federated learning of neural networks.
\newblock In {\em International Conference on Machine Learning}, pages
  7252--7261, 2019.

\bibitem{yurochkin2019statistical}
Mikhail Yurochkin, Mayank Agarwal, Soumya Ghosh, Kristjan Greenewald, and Nghia
  Hoang.
\newblock Statistical model aggregation via parameter matching.
\newblock {\em Advances in Neural Information Processing Systems},
  32:10956--10966, 2019.

\bibitem{yurochkin2018scalable}
Mikhail Yurochkin, Zhiwei Fan, Aritra Guha, Paraschos Koutris, and XuanLong
  Nguyen.
\newblock Scalable inference of topic evolution via models for latent geometric
  structures.
\newblock {\em arXiv preprint arXiv:1809.08738}, 2018.

\bibitem{wang2020federated}
Hongyi Wang, Mikhail Yurochkin, Yuekai Sun, Dimitris Papailiopoulos, and
  Yasaman Khazaeni.
\newblock Federated learning with matched averaging.
\newblock {\em arXiv preprint arXiv:2002.06440}, 2020.

\bibitem{singh2020model}
Sidak~Pal Singh and Martin Jaggi.
\newblock Model fusion via optimal transport.
\newblock {\em Advances in Neural Information Processing Systems}, 33, 2020.

\bibitem{claici2020model}
Sebastian Claici, Mikhail Yurochkin, Soumya Ghosh, and Justin Solomon.
\newblock Model fusion with kullback--leibler divergence.
\newblock {\em arXiv preprint arXiv:2007.06168}, 2020.

\bibitem{lin2020ensemble}
Tao Lin, Lingjing Kong, Sebastian~U Stich, and Martin Jaggi.
\newblock Ensemble distillation for robust model fusion in federated learning.
\newblock {\em arXiv preprint arXiv:2006.07242}, 2020.

\bibitem{zhu2021data}
Zhuangdi Zhu, Junyuan Hong, and Jiayu Zhou.
\newblock Data-free knowledge distillation for heterogeneous federated
  learning.
\newblock {\em arXiv preprint arXiv:2105.10056}, 2021.

\bibitem{chen2020fedbe}
Hong-You Chen and Wei-Lun Chao.
\newblock Fedbe: Making bayesian model ensemble applicable to federated
  learning.
\newblock {\em arXiv preprint arXiv:2009.01974}, 2020.

\bibitem{liu2022deep}
Chang Liu, Chenfei Lou, Runzhong Wang, Alan~Yuhan Xi, Li~Shen, and Junchi Yan.
\newblock Deep neural network fusion via graph matching with applications to
  model ensemble and federated learning.
\newblock In {\em International Conference on Machine Learning}, pages
  13857--13869. PMLR, 2022.

\bibitem{xiao2023probabilistic}
Peng Xiao, Biao Zhang, Samuel Cheng, Ke~Wei, and Shuqin Zhang.
\newblock Probabilistic fusion of neural networks that incorporates global
  information.
\newblock In {\em Asian Conference on Machine Learning}, pages 1149--1164.
  PMLR, 2023.

\bibitem{nguyen2021model}
Dang Nguyen, Khai Nguyen, Dinh Phung, Hung Bui, and Nhat Ho.
\newblock Model fusion of heterogeneous neural networks via cross-layer
  alignment.
\newblock {\em arXiv preprint arXiv:2110.15538}, 2021.

\bibitem{karimireddy2020scaffold}
Sai~Praneeth Karimireddy, Satyen Kale, Mehryar Mohri, Sashank Reddi, Sebastian
  Stich, and Ananda~Theertha Suresh.
\newblock Scaffold: Stochastic controlled averaging for federated learning.
\newblock In {\em International Conference on Machine Learning}, pages
  5132--5143. PMLR, 2020.

\bibitem{li2021model}
Qinbin Li, Bingsheng He, and Dawn Song.
\newblock Model-contrastive federated learning.
\newblock In {\em Proceedings of the IEEE/CVF Conference on Computer Vision and
  Pattern Recognition}, pages 10713--10722, 2021.

\bibitem{acar2021federated}
Durmus Alp~Emre Acar, Yue Zhao, Ramon~Matas Navarro, Matthew Mattina, Paul~N
  Whatmough, and Venkatesh Saligrama.
\newblock Federated learning based on dynamic regularization.
\newblock {\em arXiv preprint arXiv:2111.04263}, 2021.

\bibitem{zhang2020fedpd}
Xinwei Zhang, Mingyi Hong, Sairaj Dhople, Wotao Yin, and Yang Liu.
\newblock Fedpd: A federated learning framework with optimal rates and
  adaptivity to non-iid data.
\newblock {\em arXiv preprint arXiv:2005.11418}, 2020.

\bibitem{mohri2019agnostic}
Mehryar Mohri, Gary Sivek, and Ananda~Theertha Suresh.
\newblock Agnostic federated learning.
\newblock In {\em International Conference on Machine Learning}, pages
  4615--4625, 2019.

\bibitem{deng2021distributionally}
Yuyang Deng, Mohammad~Mahdi Kamani, and Mehrdad Mahdavi.
\newblock Distributionally robust federated averaging.
\newblock {\em arXiv preprint arXiv:2102.12660}, 2021.

\bibitem{smith2017federated}
Virginia Smith, Chao-Kai Chiang, Maziar Sanjabi, and Ameet~S Talwalkar.
\newblock Federated multi-task learning.
\newblock In {\em Advances in Neural Information Processing Systems}, pages
  4424--4434, 2017.

\bibitem{breiman1996bagging}
Leo Breiman.
\newblock Bagging predictors.
\newblock {\em Machine learning}, 24(2):123--140, 1996.

\bibitem{wolpert1992stacked}
David~H Wolpert.
\newblock Stacked generalization.
\newblock {\em Neural networks}, 5(2):241--259, 1992.

\bibitem{schapire1999brief}
Robert~E Schapire.
\newblock A brief introduction to boosting.
\newblock In {\em Ijcai}, volume~99, pages 1401--1406. Citeseer, 1999.

\bibitem{dietterich2000ensemble}
Thomas~G Dietterich.
\newblock Ensemble methods in machine learning.
\newblock In {\em International workshop on multiple classifier systems}, pages
  1--15. Springer, 2000.

\bibitem{breiman2001random}
Leo Breiman.
\newblock Random forests.
\newblock {\em Machine learning}, 45(1):5--32, 2001.

\bibitem{hinton2015distilling}
Geoffrey Hinton, Oriol Vinyals, and Jeff Dean.
\newblock Distilling the knowledge in a neural network.
\newblock {\em arXiv preprint arXiv:1503.02531}, 2015.

\bibitem{bucilua2006model}
Cristian Buciluǎ, Rich Caruana, and Alexandru Niculescu-Mizil.
\newblock Model compression.
\newblock In {\em Proceedings of the 12th ACM SIGKDD international conference
  on Knowledge discovery and data mining}, pages 535--541, 2006.

\bibitem{schmidhuber1992learning}
J{\"u}rgen Schmidhuber.
\newblock Learning complex, extended sequences using the principle of history
  compression.
\newblock {\em Neural Computation}, 4(2):234--242, 1992.

\bibitem{thibaux2007hierarchical}
Romain Thibaux and Michael~I Jordan.
\newblock Hierarchical beta processes and the indian buffet process.
\newblock In {\em Artificial Intelligence and Statistics}, pages 564--571,
  2007.

\bibitem{kairouz2021advances}
Peter Kairouz, H~Brendan McMahan, Brendan Avent, Aur{\'e}lien Bellet, Mehdi
  Bennis, Arjun~Nitin Bhagoji, Kallista Bonawitz, Zachary Charles, Graham
  Cormode, Rachel Cummings, et~al.
\newblock Advances and open problems in federated learning.
\newblock {\em Foundations and Trends{\textregistered} in Machine Learning},
  14(1--2):1--210, 2021.

\bibitem{nair2010rectified}
Vinod Nair and Geoffrey~E Hinton.
\newblock Rectified linear units improve restricted boltzmann machines.
\newblock In {\em Icml}, 2010.

\bibitem{nayak2019zero}
Gaurav~Kumar Nayak, Konda~Reddy Mopuri, Vaisakh Shaj, Venkatesh~Babu
  Radhakrishnan, and Anirban Chakraborty.
\newblock Zero-shot knowledge distillation in deep networks.
\newblock In {\em International Conference on Machine Learning}, pages
  4743--4751. PMLR, 2019.

\bibitem{nishio2019client}
Takayuki Nishio and Ryo Yonetani.
\newblock Client selection for federated learning with heterogeneous resources
  in mobile edge.
\newblock In {\em ICC 2019-2019 IEEE international conference on communications
  (ICC)}, pages 1--7. IEEE, 2019.

\bibitem{cho2020client}
Yae~Jee Cho, Jianyu Wang, and Gauri Joshi.
\newblock Client selection in federated learning: Convergence analysis and
  power-of-choice selection strategies.
\newblock {\em arXiv preprint arXiv:2010.01243}, 2020.

\bibitem{cho2022towards}
Yae~Jee Cho, Jianyu Wang, and Gauri Joshi.
\newblock Towards understanding biased client selection in federated learning.
\newblock In {\em International Conference on Artificial Intelligence and
  Statistics}, pages 10351--10375. PMLR, 2022.

\bibitem{lecun2010mnist}
Yann LeCun, Corinna Cortes, and CJ~Burges.
\newblock Mnist handwritten digit database.
\newblock {\em ATT Labs [Online]. Available: http://yann.lecun.com/exdb/mnist},
  2, 2010.

\bibitem{krizhevsky2009learning}
Alex Krizhevsky, Geoffrey Hinton, et~al.
\newblock Learning multiple layers of features from tiny images.
\newblock 2009.

\bibitem{paszke2017automatic}
Adam Paszke, Sam Gross, Soumith Chintala, Gregory Chanan, Edward Yang, Zachary
  DeVito, Zeming Lin, Alban Desmaison, Luca Antiga, and Adam Lerer.
\newblock Automatic differentiation in pytorch.
\newblock 2017.

\bibitem{kingma2014adam}
Diederik~P Kingma and Jimmy Ba.
\newblock Adam: A method for stochastic optimization.
\newblock {\em arXiv preprint arXiv:1412.6980}, 2014.

\end{thebibliography}

\end{document}